\documentclass{article}
\usepackage{spconf,amsmath,graphicx}
\usepackage{booktabs}
\usepackage{multirow}

\title{COMBINING UNSUPERVISED AND TEXT AUGMENTED SEMI-SUPERVISED LEARNING FOR LOW RESOURCED AUTOREGRESSIVE SPEECH RECOGNITION}
\name{Chak-Fai Li, Francis Keith, William Hartmann, Matthew Snover}
\address{Raytheon BBN Technologies, Cambridge MA, USA \\
{\small \tt \{chak.fai.li, francis.keith, william.hartmann, matt.snover\}@raytheon.com}}
\begin{document}
\ninept
\maketitle
\begin{abstract}
Recent advances in unsupervised representation learning have demonstrated the impact of pretraining on large amounts of read speech.
We adapt these techniques for domain adaptation in low-resource---both in terms of data and compute---conversational and broadcast domains.
Moving beyond CTC, we pretrain state-of-the-art Conformer models in an unsupervised manner.
While the unsupervised approach outperforms traditional semi-supervised training, the techniques are complementary.
Combining the techniques is a 5\% absolute improvement in WER, averaged over all conditions, compared to semi-supervised training alone.
Additional text data is incorporated through external language models.
By using CTC-based decoding, we are better able to take advantage of the additional text data.
When used as a transcription model, it allows the Conformer model to better incorporate the knowledge from the language model through semi-supervised training than shallow fusion.
Final performance is an additional 2\% better absolute when using CTC-based decoding for semi-supervised training compared to shallow fusion.
\end{abstract}
\section{Introduction}
\label{sec:intro}

Epistemic uncertainty \cite{hullermeier2019aleatoric}, due to limited training data, is a universal problem for automatic speech recognition (ASR), and machine learning in general.
Transcribed data is difficult to obtain because it is either expensive or restricted due to privacy or proprietary reasons.
Some companies do have access to tremendous amounts of transcribed data, but it cannot be shared beyond the company.
It is also possible to obtain large amounts of transcribed data in certain domains like read speech \cite{panayotov2015librispeech}, but that data is unlikely to be useful in building models for more difficult domains like far-field and conversational speech.
While gathering large quantities of labeled data is difficult, unlabeled data is easier to find.
Assuming the model's purpose is to process large amounts of data, then this same data can serve as untranscribed training data.
Even if a human is not allowed to inspect the data for privacy or security reasons, an automated system may still use the data for training.

A common approach to incorporating untranscibed data in ASR training is through semi-supervised training (SST)---the use of pseudotranscripts from an initial model for supervision.
A strong lexicon and language model---provided externally to a hybrid model or implicit in a sequence-to-sequence model---restricts the output to sequences of likely words in the language.
Because of the extra information added by the language model, SST works remarkably well for ASR.
When trying to improve a well-trained supervised model, it can take an order of magnitude more unsupervised data to provide an improvement \cite{huang2013semi, parthasarathi2019lessons}.
In the case of domain adaptation, SST can produce large gains with just a few hours of data \cite{wotherspoon2021improved}.

While SST performs well for ASR, it does have drawbacks.
The output from the transcription model is a discrete word sequence that has removed uncertainty in the output.
Alternatives that utilize the lattice have been proposed, but additional gains are minimal \cite{manohar2018semi}.
Most work uses academic datasets where the untranscribed data has already been filtered.
We can assume the data is of high quality and matched to the domain of interest.
With wild data, the data itself is questionable.
Portions of the data may be from the wrong domain, wrong language, or even contain no discernible speech at all.
Using wild data indiscriminately can harm the final model. 
Even in cases where the data is curated, large errors in the hypothesized transcripts, especially in the form of deletions, can harm the final model \cite{wotherspoon2021improved}.

An alternative, and potentially complementary approach is unsupervised learning.
Unsupervised representation learning can address some of the concerns with SST.
While SST requires curation of the untranscribed data, unsupervised training should be  more robust to low quality, or out-of-domain data.
Since we are not assigning real labels, poor hypothesized transcripts cannot corrupt the model.
The model might still waste capacity modeling irrelevant data, but that is less detrimental than hallucinating words for out-of-domain data.

Unsupervised learning approaches work by defining a \emph{pretext} task. 
We can largely separate the approaches into two categories, contrastive and reconstruction-based.
The fundamental difference between the two approaches is that the contrastive approaches require negative samples.
The model generates a candidate output and compares it with the true target and a number of negative samples.
The objective function is based on the candidate's similarity with the target and distance from the negative samples.
In \cite{vandenoord2018representing}, future states are predicted from the current state.
The more recent \emph{wav2vec 2.0} approach predicts a state given surrounding context \cite{baevski2020wav2vec}, similar to the way BERT \cite{devlin2018bert} is trained for NLP tasks.

In reconstruction-based approaches have no negative samples; the model directly attempts to reconstruct the input.
Autoencoder-style approaches are typical reconstruction-based approaches \cite{neumann2019improving, liu2021tera, chung2019unsupervised}.
A variant of wav2vec 2.0, DeCoAR 2.0 \cite{ling2020decoar} replaces the contrastive loss with a reconstruction loss.
The recent HuBERT \cite{hsu2021hubert} approach can also be considered a reconstruction-based approach.
We discuss the HuBERT approach in more detail in the next section.

Reconstruction has the potential downside that the model has no prior knowledge of what parts of the representation are more important than others, so all errors are weighted equally.
When dealing with high-dimensional signals (e.g. speech or images), modeling the underlying relationships between the dimensions in the original feature space is difficult and costly.
Reconstructing every detail is likely unnecessary as there are many nuisance dimensions that are irrelevant for the eventual task.
We know the number of relevant classes/categories are orders of magnitude smaller than the number of unique signals.
We also know the objective function cannot distinguish between irrelevant differences in the reconstructed signal.
Approaches like CPC \cite{vandenoord2018representing}, wav2vec 2.0 \cite{baevski2020wav2vec},  DeCoAR 2.0 \cite{ling2020decoar} use quantized or latent representations representations to allow the model to focus on the more relevant dimensions during reconstruction.

In this work we use the HuBERT approach to unsupervised learning to pretrain the encoder of a Conformer model for the purposes of domain adaptation.
The model is then fine-tuned on out-of-domain supervised data.
We compare unsupervised learning with semi-supervised learning, both separately and in combination.
We also have access to large amounts of text data in the new domain. 
Despite the advance of techniques like shallow fusion, deep fusion \cite{gulcehre2015using}, cold fusion \cite{sriram2017cold}, and internal language model estimation \cite{meng2021internal}, there is a limit to how much an autoregressive model can gain from additional language model data.
We experiment with utilizing the encoder from the Conformer model as a CTC model in a non-autoregressive framework in order to take full advantage of the additional text data.
Our contributions include:

\begin{itemize}
    \item Application of unsupervised learning in a low-resource domain adaptation task with an autoregressive model
    \item Integration of external language model information through semi-supervised training
    \item Confirmation that unsupervised pretraining is both more powerful and complementary with semi-supervised learning.
\end{itemize}

In Section \ref{sec:method} we describe our approaches to unsupervised and semi-supervised training. Section \ref{sec:setup} details the data and experimental setup. Our results are presented and discussed in Section \ref{sec:results}, and conclusions are presented in Section \ref{sec:conclusion}.

\section{Learning from Untranscribed Data}\label{sec:method}

\subsection{Unsupervised Pretraining Approach}

The structure of the network in the HuBERT approach \cite{hsu2021hubert} is similar to wav2vec 2.0 \cite{baevski2020wav2vec}.
Both use the same convolutional waveform encoder followed by a large transformer network, however, HuBERT does not use a quantization component.
The first step in the training process requires a k-means clustering of the features to generate unsupervised targets.
The initial features are MFCC features, but they are used to bootstrap the initial model and later iterations use an internal representation from the transformer network. 
Given the clusters as targets, the model is trained using frame-level cross-entropy.
Learning the cluster targets would be trivial for a sophisticated model.
To combat a degenerative solution, some of the inputs to the transformer network are masked and only updates related to the masked frames are used; instead of directly predicting the target from the input, the model predicts the target based on context.

While we are inspired by the HuBERT approach \cite{hsu2021hubert}, our implementation and use differs in several aspects from the original work.
Our design decisions stem from a desire to reduce computational and data requirements.
Instead of building the initial clusters from MFCC features, we first train a Conformer \cite{gulati2020conformer} model on a small amount of supervised out-of-domain data.
Once the model has been trained, the decoder portion of the network is discarded and only the encoder is kept.
We use the final layer of the encoder to generate embeddings for the untranscribed data.
After embedding the acoustic signal with the encoder, we cluster the embedded features using k-means.
As in \cite{hsu2021hubert}, the clusters serve as labels for the unsupervised training.
We remove the convolutional waveform encoder and train a Conformer encoder from filterbank features instead.
During this step, there is no decoder involved.
The model is trained using frame-level cross-entropy using the clusters as targets.
After the encoder has been pretrained using the untranscribed data, a randomly initialized decoder is appended and the entire model is fine-tuned using the original transcribed audio.

Our work also differs from HuBERT in terms of the task.
Our data consists of a mix of CTS and broadcast news, more difficult domains than read speech.
The HuBERT paper used a minimum of 960 hours and as much as 60k hours, much more data than we have available.
We consider the task of domain adaptation with unsupervised learning.
The models trained in the HuBERT paper were CTC models with external LMs.
We focused on encoder-decoder style models, both with and without external LMs.
The HuBERT paper mentions that the number of GPUs used during unsupervised pretraining is critical because of batch size.
A minimum of 16 GPUs was required to achieve good results.
We found performance to be stable with respect to the number of GPUs used, with a single GPU being sufficient.
The original HuBERT paper focuses only on masked frames during unsupervised learning.
We perform the same update on the model regardless of whether the original frame was masked.

We do not report these results, but our model failed to improve over the baseline when the clusters were generated using MFCC features.
While best performance was seen by clustering features extracted from an encoder, the original HuBERT paper was able to learn using the MFCC features for clustering.
The reason for our failure to learn using MFCC-based clusters could be due to a variety of factors including more difficult, limited data, and limited GPUs.

\subsection{Semi-Supervised Learning}

The standard approach to semi-supervised training uses an initial model to transcribe the unlabeled data and then treats the hypothesized transcripts as truth.
We start with the supervised model as our transcription model.
During SST we use both the supervised data and the untranscribed data, making no distinction between the two during the training process.
For hybrid models keeping only a subset of the hypothesized transcripts can be important.
The common approach is to simply filter by confidence, but more sophisticated methods are sometimes used \cite{wotherspoon2021improved, de2016high}.
For autoregressive models, we found selection to be less of an issue.
The models are data hungry and tend to perform better with more data, even if it is lower quality.

The recent \emph{noisy student} approach \cite{park2020improved, zhang2020pushing} is also related.
That approach extends the classical approach to SST by applying more data augmentation and a more careful filtering of the pseudotranscripts.
Typically the noisy student work also increases the model size as the amount of data increases.
Since the untranscribed data in our sets is significantly less than the amount used in the noisy student work, we do not apply these additional improvements.

\section{Experimental Setup}\label{sec:setup}

\subsection{MATERIAL Data}

\begin{table}
    \caption{\label{tab:data} {\it Amount of transcribed CS and untranscribed CS and BN data for each language. Most of the untrascribed data is BN and no transcribed BN data is available.}}
    \centering
    \begin{tabular}{lrrrr}
    \toprule
        \multirow{2}{*}{Language} & \multicolumn{2}{c}{Transcribed} & \multicolumn{2}{c}{Untranscribed}\\
        \cmidrule(lr){2-3} \cmidrule(lr){4-5}
        & CS & BN & CS & BN \\
        \midrule
        Bulgarian & 41.1 & 0 & 33.3 & 149.9 \\
        Swahili & 68.3 & 0 & 57.6 & 149.0 \\ 
        Tagalog & 127.9 & 0 & 48.4 & 153.8 \\
    \bottomrule
    \end{tabular}
\end{table}

We use three languages from the IARPA MATERIAL\footnote{https://www.iarpa.gov/index.php/research-programs/material} program: Bulgarian, Swahili, and Tagalog.
The languages are a representative sample of the nine languages from the program in terms of difficulty.
All transcribed training data in the program consists of conversational telephone speech (CS).
The test data contains a small amount of CS data, but mostly consists of news broadcast (NB) and topical broadcast (TB) data.
The same is true of the additional untranscribed data used for training.
For the remainder of the paper, we combine the two broadcast sets into a single set labeled broadcast (BN).
This represents a large domain shift between the training and test data.
In addition to the domain shift, the out-of-vocabulary (OOV) rate is high for all languages.
To address the domain shift from the text side we augment that text data with source-side parallel data provided by the MATERIAL program, ParaCrawl \cite{espla2019paracrawl} data (where available), and automatically collected web data \cite{zhang2015enhancing}.
The total amount of text data ranges from 80 to 120 million words per language.
Note that the text data is still out-of-domain, though likely a closer match than the original acoustic transcripts.
The distribution of domain for the acoustic data is shown in Table \ref{tab:data}.
For more details about the data, see \cite{li2021overcoming}.
Note that the transcribed CS data for Swahili and Tagalog is identical to the data from the IARPA BABEL program, available through the LDC (LDC2017S05, LDC2016S13).

\subsection{Conformer Model Training}

Our sequence-to-sequence models are conformer-based \cite{gulati2020conformer} encoder-decoder models trained in EspNet \cite{watanabe2018espnet}.
The configuration is similar to the ones described in \cite{guo2021recent}.
The encoder has 12 layers with four attention heads, an embedding dimension of 256, and a FFN dimension of 2048.
The decoder uses 6 layers with identical parameters.
In addition to the standard cross-entropy objective function, we also use the CTC objective function \cite{kim2017joint}.
Our output units are characters.

During unsupervised pretraining, we use frame-level cross entropy to train the Conformer encoder with the k-means clusters as targets.
The encoder is trained for a maximum of 40 epochs.
The minibatch size is 128 utterances, corresponding to approximately 500 seconds of audio.
We use SpecAugment \cite{park2019specaugment} during both supervised and unsupervised training.

\subsection{Combining the Encoder with N-Gram Language Models}

Since our Conformer models are jointly trained with a cross-entropy and CTC objective function, the encoder can be used separately from the decoder for ASR.
Using the encoder alone makes the model non-autoregressive.
One benefit of a non-autoregressive model is it will generate posteriors for each output unit at each frame, making it easy to combine the model with an external word-level LM and lexicon.

We combine a weighted finite state transducer (WFST)-based decoder with the Conformer encoder in order to utilize an expanded lexicon and n-gram language model. For our experiments, we use trigram LMs.
This has been done before in \cite{miao2015eesen}\cite{zenkel2017comparison}, and in particular our approach is similar to WFST decoding in \cite{zenkel2017comparison} using a decoder implemented in Kaldi \cite{povey2011kaldi}.
The WFST is a simple composition of three transducers: a token transducer, which removes blank symbols output by CTC and collapses repeated characters; a lexicon transducer, which maps sequences of collapsed characters into words; and a grammar transducer, which contains the n-gram LM. 
%Utilizing a CTC-based model in a WFST-decoder is commonly done in the literature \cite{miao2015eesen, zenkel2017comparison}.

\section{Results}\label{sec:results}

\subsection{Impact of Cluster Type in Unsupervised Training}

\begin{table}
    \caption{\label{tab:cluster} {\it Performance (WER) of unsupervised clusters from K-Means compared to using position-dependent phone labels.}}
    \centering
    \begin{tabular}{llcc}
    \toprule
        Language & Cluster Type & CS & BN \\
        \midrule
        Bulgarian & Position Dependent Phones & 40.1 & 38.7 \\
        %Bulgarian & 200 & 36.0 & 40.7 \\ 
        %Bulgarian & 500 & 34.5 & 40.3 \\ 
        %Bulgarian & 1000 & 35.2 & 41.0 \\ 
        Bulgarian & K-Means (5000)  & 35.0 & 40.2 \\
        %Bulgarian & 10000 & 34.4 & 41.9 \\ 
        \midrule
        Swahili & Position Dependent Phones & 38.1 & 57.9 \\
        Swahili & K-Means (5000) & 33.7 & 49.0 \\ 
        \midrule
        Tagalog & Position Dependent Phones & 44.8 & 52.8 \\
        Tagalog & K-Means (5000) & 38.8 & 53.3 \\ 
    \bottomrule
    \end{tabular}
\end{table}

For all three languages we compared the unsupervised targets with using position-dependent phones as targets.
Results are in Table \ref{tab:cluster}.
Results are similar to using the unsupervised clusters, except for Swahili where the unsupervised clusters are significantly better.
Note that we do not have ground truth transcripts for the unsupervised data, so the phone targets come from forced alignment with the hypothesized transcripts from a hybrid model.
We did tune the number of clusters on Bulgarian, cluster size has little impact on overall performance.
In all cases we are using the output from a supervised encoder for clustering.
We were unable to improve over a baseline model when using MFCC features for clustering.

\subsection{Performance without Additional LM Data}

\begin{table}
    \setlength{\tabcolsep}{5pt}
    \caption{\label{tab:nolm} {\it Comparing WER performance using unsupervised and semi-supervised training without external language model data.}}
    \centering
    \begin{tabular}{@{}lcccccc@{}}
    \toprule
        \multirow{2}{*}{Model Description} & \multicolumn{2}{c}{Bulgarian} & \multicolumn{2}{c}{Swahili} & \multicolumn{2}{c}{Tagalog}\\
        \cmidrule(lr){2-3} \cmidrule(lr){4-5} \cmidrule(lr){6-7}
        & CS & BN & CS & BN & CS & BN \\
        \midrule
        Supervised & 39.1 & 50.7 & 37.5 & 58.1 & 43.5 & 65.0 \\ 
        \midrule
        SST iter. 1 & 35.5 & 41.5 & 35.7 & 49.7 & 42.4 & 57.6 \\
        SST iter. 2 & 34.8 & 36.9 &  36.3 &  47.0 & 42.6 & 53.7  \\ 
        \midrule
        Unsup iter. 1 & 35.0 & 40.2 & 33.7 & 49.0 & 38.8 & 53.3 \\
        Unsup iter. 2 & 32.1 & 35.0 & 33.6 & 44.2 & \textbf{38.6} & 49.4 \\
        \midrule
        SST from Unsup iter. 1 & \textbf{31.9} & \textbf{33.4} & \textbf{33.4} & \textbf{36.9} & 39.2 & \textbf{47.8} \\ 
    \bottomrule
    \end{tabular}
\end{table}

Results without an external language model can be seen in Table \ref{tab:nolm}.
All models are Conformer encoder-decoder models; the external LM is a two-layer LSTM model.
The supervised model, trained only on CS, sets the baseline for all subsequent models. 
While WER for BN is typically lower than CS in most languages, that is not that case in our results.
The supervised model has never seen broadcast data and the BN data has a high OOV rate, explaining the discrepancy.

We directly compare the WER reduction from semi-supervised and unsupervised training. 
The semi-supervised model shows a clear improvement across all conditions, but the unsupervised model consistently outperforms the semi-supervised model (41.7\% average WER vs. 43.7\%).
When considering multiple iterations, two rounds of unsupervised training is better than two rounds of SST. In fact, the gap between the unsupervised and semi-supervised approach grows from 2.0\% absolute to 3.1\%.
We also consider combining the two approaches by applying SST using the initial unsupervised model for transcription.
Combining the two approaches gives the best overall performance---4.8\% absolute improvement compared to multiple iterations of SST alone---demonstrating the complementarity of the unsupervised and semi-supervised approaches.

The overall improvement on BN data is not surprising given the lack of BN data in the supervised model, but the gain in CS is also significant.
Unsupervised pretraining improves the CS result of Tagalog by almost five points absolute.
Tagalog had the largest amount of supervised CS data ($\approx$ 120 hours) and there was only an additional 50 hours in the untranscribed set.
The unsupervised pretraining clearly provides benefits for both in and out-of-domain data.

\subsection{Performance with additional LM data}

\begin{table}[ht]
    \setlength{\tabcolsep}{3pt}
    \caption{\label{tab:withlm} {\it Comparing performance using supervised, unsupervised, and semi-supervised training with external language data. For each model comparisons are given using shallow fusion and CTC decoding with an N-Gram LM}}
    \centering
    \begin{tabular}{@{}llcccccc@{}}
    \toprule
        \multirow{2}{*}{\textbf{Supervision}} & \multirow{2}{*}{\textbf{Transcription}} & \multicolumn{2}{c}{Bulgarian} & \multicolumn{2}{c}{Swahili} & \multicolumn{2}{c}{Tagalog}\\
        \cmidrule(lr){3-4} \cmidrule(lr){5-6} \cmidrule(lr){7-8}
        & \textbf{Model} & CS & BN & CS & BN & CS & BN \\
        \midrule
        \textbf{Supervised} & \textbf{none} &  &  &  &  &  &  \\
        \phantom{ab} Fusion &  & 37.6 & 41.8 & 38.6 & 52.6 & 46.0 & 62.5 \\
        \phantom{ab} CTC &  & 39.2 & 36.4 & 38.6 & 48.1 & 46.5 & 56.8 \\
        %\phantom{ab} CTC Big &  & 38.1 & 33.5 & 38.6 & 47.6 & 46.6 & 55.2 \\
        \midrule
        \textbf{Unsup.} & \textbf{none} &  &  &  &  &  &  \\
        \phantom{ab} Fusion & & 35.1 & 32.9 & 35.6 & 44.2 & 41.4 & 51.1  \\ 
        \phantom{ab} CTC & & 37.5 & 29.1 & 36.4 & 40.0 & 42.0 & 48.7  \\
        %\phantom{ab} CTC Big & & 36.3 & 26.4 & 36.6 & 39.1 & 42.1 & 47.1  \\
        \midrule
        \textbf{SST} & \textbf{Sup+Fusion} &  &  &  &  &  &  \\
        \phantom{ab} Fusion & & 33.7 & 32.2 & 37.3 & 43.0 & 43.4 & 55.4  \\ 
        \phantom{ab} CTC &  & 38.4 & 30.9 & 39.0 & 39.3 & 45.3 & 49.6  \\
        %\phantom{ab} CTC Big &  & 37.5 & 28.0 & 39.0 & 38.6 & 45.2 & 48.2  \\ 
        \midrule
        \textbf{SST} & \textbf{Unsup+Fusion} &  &  &  &  &  &  \\
        \phantom{ab} Fusion &  & 31.6 & 26.4 & 35.3 & 38.3 & 38.6 & 44.9  \\
        \phantom{ab} CTC &  & 37.5 & 26.9 & 36.3 & 35.6 & 41.2 & 41.8  \\
        %\phantom{ab} CTC Big &  & 36.4 & 24.6 & 36.2 & 35.0 & 41.6 & 39.9  \\ 
        \midrule
        \textbf{SST} & \textbf{Unsup+CTC} &  &  &  &  &  &  \\
        \phantom{ab} Fusion & & \textbf{30.8} & \textbf{24.5} & \textbf{33.4} & \textbf{33.7} & \textbf{38.2} & \textbf{41.0}  \\
        \phantom{ab} CTC & & 36.1 & 27.3 & 35.1 & 34.9 & 41.8 & 44.1  \\ 
        %\phantom{ab} CTC Big & & 34.7 & 25.6 & 35.2 & 34.5 & 42.0 & 42.9  \\ 
        %\midrule
        %\textbf{SST} & \textbf{Unsup+CTC-big} &  &  &  &  &  &  \\ 
        %\phantom{ab} Fusion & & 30.5 & 22.5 & 33.0 & 33.3 & 39.4 & 39.9  \\ 
        %\phantom{ab} CTC & & 35.6 & 25.7 & 35.0 & 34.5 & 42.1 & 43.1 \\ 
        %\phantom{ab} CTC Big & & 34.4 & 23.9 & 35.2 & 34.2 & 42.0 & 42.1 \\
    \bottomrule
    \end{tabular}
\end{table}

Table \ref{tab:withlm} contains the full set of results comparing unsupervised and SST with the use of external LMs.
Initially, we consider the use of shallow fusion and ignore the CTC results.
Overall, we can see a dramatic improvement in the performance on BN for all languages and models.
The external LM has a much lower OOV rate and the additional language data is more similar to BN data than CS data.
Once the external LM has been incorporated, the superiority of the unsupervised approach over the SST approach is no longer seen; they each outperform the other half the time.
However, best performance is still achieved by combining the two approaches.
Average performance of the supervised model is 46.5\%.
A single round of SST improves the WER to 40.8\%, while SST on top of the unsupervised pretraining further improves the WER to 35.9\%.

\subsection{CTC-based Decoding with N-Gram LMs}

In addition to shallow fusion, Table \ref{tab:withlm} contains results using the CTC encoder with an external lexicon and LM.
Entries marked with \emph{CTC} use a trigram LM trained on the same data as the model used for shallow fusion.
%As it is cheap and easy to add additional training data for an N-Gram language model, we also experiment with using a significantly larger set of text data (\emph{CTC Big N-Gram}). 
%We also experiment with using a significantly larger set of text data (\emph{CTC Big N-Gram}).
%This data is trivial to add to an n-gram language model, but is problematically large for tuning external recurrent models for shallow fusion. As such, we only report the results on CTC decoding with this set.
CTC only decoding is significantly worse than using the full Conformer model.
Due to space, full results are not shown, but CTC decoding is about 5 to 8 percent absolute worse than the Conformer model when no external LM is used for either.
However, CTC decoding is able to recover the gap and improve upon the Conformer model through the use of an external lexicon and LM.
For the supervised, unsupervised, and semi-supervised models (using a transcription model with shallow fusion), the CTC decoding outperforms the full Conformer model with shallow fusion on BN data.
For example, the supervised model is improved by 4.5\% to 5.7\% absolute on BN data when using CTC decoding.
Performance on CS data is worse though, likely due the external lexicon and LM providing less additional information for the CS data.

%However, CTC decoding is able to recover the gap and improve upon the Conformer model through the use of an external lexicon and LM.
%For the supervised, unsupervised, and semi-supervised models (using a transcription model with shallow fusion), the CTC decoding outperforms the full Conformer model with shallow fusion on BN data.
%Performance on CS data is worse though.
%Presumably the external lexicon and LM are providing less additional information for the CS data.

The conclusion from the final semi-supervised model, where the transcription comes from a CTC decode, shows a different pattern.
Even for the BN domain, CTC decoding no longer provides an improvement over shallow fusion.
When the transcription model uses CTC decoding, the resulting semi-supervised model will have largely incorporated that information into the model.
The additional information in the lexicon and LM is no longer beneficial.
The CTC decoding is helpful, but only for the initial transcription.
Using CTC decoding for transcription in semi-supervised training yields an average WER of 33.6\% across all conditions when the full Conformer model is used for decoding with shallow fusion, a 2.3\% average improvement over using shallow fusion for pseudotranscription.
%Once it has been used once, it can be discarded.

%Finally, we examine the use of CTC decoding as a transcription model in the final two sets of semi-supervised models, which shows a very different pattern.
%Even using the larger N-gram LM, CTC decoding no longer yields an improvement over shallow fusion on the BN subset.
%This shows that if the transcription model uses CTC decoding with an external language model, the resulting semi-supervised model is able to incorporate that information into the Conformer model.
%While continuing to use CTC decoding is no longer useful, the resulting Conformer models trained on the transcripts generated with CTC decoding yield our best results on both the CS and BN subsets.

\section{Conclusions}\label{sec:conclusion}

Unsupervised representation learning is a powerful approach to using untranscribed data in an ASR system.
Prior work has focused on large amounts of read speech.
We demonstrate the applicability to smaller amounts of more difficult data---CTS and BN.
The approach works on a single GPU and does not require multiple GPUs to increase the minibatch size.
Previous work has focused on non-autoregressive models trained with CTC, but we show the approach can improve an autoregressive Conformer model.
In a direct comparison with SST approaches, the unsupervised approach proves more effective; though we achieve our best performance by combining the techniques---5\% absolute better than SST alone.

We demonstrated that external language models can significantly improve performance on out-of-domain test sets.
While an autoregressive model like the Conformer model is generally superior to a non-autoregressive CTC approach, we show that CTC decoding can make more effective use of external language model data.
By using CTC decoding with an external LM as a transcription model for semi-supervised training, we are able to incorporate the information into the Conformer model, leading to an overall additional improvement of 2\% absolute.
Once we have used CTC decoding in semi-supervised transcription, it becomes better to use the full autoregressive semi-supervised model with shallow fusion for decoding compared to decoding with CTC.
%However, once the additional information, has been incorporated into the model through semi-supervised training, CTC decoding no longer helps.

We plan to continue to improve our unsupervised learning approach.
While semi-supervised training is mature, unsupervised learning is a less explored space.
We expect further optimization will lead to even larger gains.

\section{ACKNOWLEDGEMENTS}

This work was supported by the Intelligence Advanced Research Projects Activity (IARPA) via Department of Defense US Air Force Research Laboratory contract number FA8650-17-C-9118.
This document does not contain technology or Technical Data controlled under either the U.S. International Traffic in Arms Regulations or the U.S. Export Administration Regulations.

\vfill\pagebreak

\bibliographystyle{IEEEbib}
\bibliography{refs}

\begin{thebibliography}{10}

\bibitem{hullermeier2019aleatoric}
Eyke H{\"u}llermeier and Willem Waegeman,
\newblock ``Aleatoric and epistemic uncertainty in machine learning: A tutorial
  introduction,''
\newblock {\em arXiv preprint arXiv:1910.09457}, 2019.

\bibitem{panayotov2015librispeech}
Vassil Panayotov, Guoguo Chen, Daniel Povey, and Sanjeev Khudanpur,
\newblock ``Librispeech: an asr corpus based on public domain audio books,''
\newblock in {\em IEEE ICASSP}, 2015, pp. 5206--5210.

\bibitem{huang2013semi}
Yan Huang, Dong Yu, Yifan Gong, and Chaojun Liu,
\newblock ``Semi-supervised gmm and dnn acoustic model training with
  multi-system combination and confidence re-calibration.,''
\newblock in {\em Interspeech}, 2013, pp. 2360--2364.

\bibitem{parthasarathi2019lessons}
Sree Hari~Krishnan Parthasarathi and Nikko Strom,
\newblock ``Lessons from building acoustic models with a million hours of
  speech,''
\newblock in {\em IEEE ICASSP}, 2019, pp. 6670--6674.

\bibitem{wotherspoon2021improved}
Shannon Wotherspoon, William Hartmann, Matthew Snover, and Owen Kimball,
\newblock ``Improved data selection for domain adaptation in asr,''
\newblock in {\em IEEE ICASSP}, 2021.

\bibitem{manohar2018semi}
Vimal Manohar, Hossein Hadian, Daniel Povey, and Sanjeev Khudanpur,
\newblock ``Semi-supervised training of acoustic models using lattice-free
  mmi,''
\newblock in {\em IEEE ICASSP}, 2018, pp. 4844--4848.

\bibitem{vandenoord2018representing}
A{\"{a}}ron van~den Oord, Yazhe Li, and Oriol Vinyals,
\newblock ``Representation learning with contrastive predictive coding,'' 2018.

\bibitem{baevski2020wav2vec}
Alexei Baevski, Henry Zhou, Abdelrahman Mohamed, and Michael Auli,
\newblock ``wav2vec 2.0: A framework for self-supervised learning of speech
  representations,'' 2020.

\bibitem{devlin2018bert}
Jacob Devlin, Ming-Wei Chang, Kenton Lee, and Kristina Toutanova,
\newblock ``Bert: Pre-training of deep bidirectional transformers for language
  understanding,''
\newblock {\em arXiv preprint arXiv:1810.04805}, 2018.

\bibitem{neumann2019improving}
Michael Neumann and Ngoc~Thang Vu,
\newblock ``Improving speech emotion recognition with unsupervised
  representation learning on unlabeled speech,''
\newblock in {\em IEEE ICASSP}, 2019, pp. 7390--7394.

\bibitem{liu2021tera}
Andy~T Liu, Shang-Wen Li, and Hung-yi Lee,
\newblock ``Tera: Self-supervised learning of transformer encoder
  representation for speech,''
\newblock {\em IEEE/ACM TASLP}, vol. 29, pp. 2351--2366, 2021.

\bibitem{chung2019unsupervised}
Yu-An Chung, Wei-Ning Hsu, Hao Tang, and James Glass,
\newblock ``An unsupervised autoregressive model for speech representation
  learning,''
\newblock {\em arXiv preprint arXiv:1904.03240}, 2019.

\bibitem{ling2020decoar}
Shaoshi Ling and Yuzong Liu,
\newblock ``Decoar 2.0: Deep contextualized acoustic representations with
  vector quantization,''
\newblock {\em arXiv preprint arXiv:2012.06659}, 2020.

\bibitem{hsu2021hubert}
Wei-Ning Hsu, Benjamin Bolte, Yao-Hung~Hubert Tsai, Kushal Lakhotia, Ruslan
  Salakhutdinov, and Abdelrahman Mohamed,
\newblock ``Hubert: Self-supervised speech representation learning by masked
  prediction of hidden units,''
\newblock {\em arXiv preprint arXiv:2106.07447}, 2021.

\bibitem{gulcehre2015using}
Caglar Gulcehre, Orhan Firat, Kelvin Xu, Kyunghyun Cho, Loic Barrault, Huei-Chi
  Lin, Fethi Bougares, Holger Schwenk, and Yoshua Bengio,
\newblock ``On using monolingual corpora in neural machine translation,''
\newblock {\em arXiv preprint arXiv:1503.03535}, 2015.

\bibitem{sriram2017cold}
Anuroop Sriram, Heewoo Jun, Sanjeev Satheesh, and Adam Coates,
\newblock ``Cold fusion: Training seq2seq models together with language
  models,''
\newblock {\em arXiv preprint arXiv:1708.06426}, 2017.

\bibitem{meng2021internal}
Zhong Meng, Sarangarajan Parthasarathy, Eric Sun, Yashesh Gaur, Naoyuki Kanda,
  Liang Lu, Xie Chen, Rui Zhao, Jinyu Li, and Yifan Gong,
\newblock ``Internal language model estimation for domain-adaptive end-to-end
  speech recognition,''
\newblock in {\em IEEE SLT}, 2021, pp. 243--250.

\bibitem{gulati2020conformer}
Anmol Gulati, James Qin, Chung-Cheng Chiu, Niki Parmar, Yu~Zhang, Jiahui Yu,
  Wei Han, Shibo Wang, Zhengdong Zhang, Yonghui Wu, and R.~Pang,
\newblock ``Conformer: Convolution-augmented transformer for speech
  recognition,'' 2020.

\bibitem{de2016high}
F{\'e}lix de~Chaumont~Quitry, Asa Oines, Pedro Moreno, and Eugene Weinstein,
\newblock ``High quality agreement-based semi-supervised training data for
  acoustic modeling,''
\newblock in {\em IEEE SLT}, 2016, pp. 592--596.

\bibitem{park2020improved}
Daniel~S. Park, Yu~Zhang, Ye~Jia, Wei Han, Chung-Cheng Chiu, Bo~Li, Yonghui Wu,
  and Quoc~V. Le,
\newblock ``Improved noisy student training for automatic speech recognition,''
  2020.

\bibitem{zhang2020pushing}
Yu~Zhang, James Qin, Daniel~S Park, Wei Han, Chung-Cheng Chiu, Ruoming Pang,
  Quoc~V Le, and Yonghui Wu,
\newblock ``Pushing the limits of semi-supervised learning for automatic speech
  recognition,''
\newblock {\em arXiv preprint arXiv:2010.10504}, 2020.

\bibitem{espla2019paracrawl}
Miquel Espl{\`a}-Gomis, Mikel~L Forcada, Gema Ram{\'\i}rez-S{\'a}nchez, and
  Hieu Hoang,
\newblock ``Paracrawl: Web-scale parallel corpora for the languages of the
  eu,''
\newblock in {\em Proceedings of Machine Translation Summit XVII}, 2019, pp.
  118--119.

\bibitem{zhang2015enhancing}
Le~Zhang, Damianos Karakos, William Hartmann, Roger Hsiao, Richard Schwartz,
  and Stavros Tsakalidis,
\newblock ``Enhancing low resource keyword spotting with automatically
  retrieved web documents,''
\newblock in {\em Interspeech}, 2015.

\bibitem{li2021overcoming}
Chak-Fai Li, Francis Keith, William Hartmann, Matthew Snover, and Owen Kimball,
\newblock ``Overcoming domain mismatch in low resource sequence-to-sequence asr
  models using hybrid generated pseudotranscripts,''
\newblock {\em arXiv preprint arXiv:2106.07716}, 2021.

\bibitem{watanabe2018espnet}
Shinji Watanabe, Takaaki Hori, Shigeki Karita, Tomoki Hayashi, Jiro Nishitoba,
  Yuya Unno, Nelson Enrique~Yalta Soplin, Jahn Heymann, Matthew Wiesner, Nanxin
  Chen, et~al.,
\newblock ``Espnet: End-to-end speech processing toolkit,''
\newblock {\em arXiv preprint arXiv:1804.00015}, 2018.

\bibitem{guo2021recent}
Pengcheng Guo, Florian Boyer, Xuankai Chang, Tomoki Hayashi, Yosuke Higuchi,
  Hirofumi Inaguma, Naoyuki Kamo, Chenda Li, Daniel Garcia-Romero, Jiatong Shi,
  et~al.,
\newblock ``Recent developments on espnet toolkit boosted by conformer,''
\newblock in {\em IEEE ICASSP}, 2021, pp. 5874--5878.

\bibitem{kim2017joint}
Suyoun Kim, Takaaki Hori, and Shinji Watanabe,
\newblock ``Joint ctc-attention based end-to-end speech recognition using
  multi-task learning,''
\newblock in {\em IEEE ICASSP}, 2017, pp. 4835--4839.

\bibitem{park2019specaugment}
Daniel~S Park, William Chan, Yu~Zhang, Chung-Cheng Chiu, Barret Zoph, Ekin~D
  Cubuk, and Quoc~V Le,
\newblock ``Specaugment: A simple data augmentation method for automatic speech
  recognition,''
\newblock {\em arXiv preprint arXiv:1904.08779}, 2019.

\bibitem{miao2015eesen}
Yajie Miao, Mohammad Gowayyed, and Florian Metze,
\newblock ``Eesen: End-to-end speech recognition using deep rnn models and
  wfst-based decoding,''
\newblock in {\em IEEE ASRU}, 2015, pp. 167--174.

\bibitem{zenkel2017comparison}
Thomas Zenkel, Ramon Sanabria, Florian Metze, Jan Niehues, Matthias Sperber,
  Sebastian St{\"u}ker, and Alex Waibel,
\newblock ``Comparison of decoding strategies for ctc acoustic models,''
\newblock {\em arXiv preprint arXiv:1708.04469}, 2017.

\bibitem{povey2011kaldi}
Daniel Povey, Arnab Ghoshal, Gilles Boulianne, Lukas Burget, Ondrej Glembek,
  Nagendra Goel, Mirko Hannemann, Petr Motlicek, Yanmin Qian, Petr Schwarz,
  et~al.,
\newblock ``The kaldi speech recognition toolkit,''
\newblock in {\em IEEE ASRU}, 2011.

\end{thebibliography}

\end{document}